\documentclass[a4paper, 10pt, conference]{IEEEtran}
\IEEEoverridecommandlockouts
\usepackage{cite}
\usepackage{amsmath,amssymb,amsfonts}
\usepackage{algorithmic}
\usepackage{graphicx}
\usepackage{textcomp}
\usepackage{xcolor}
\usepackage{mathtools}
\usepackage{amsthm}
\usepackage{tabularx}
\usepackage{listings}
\newtheorem{theorem}{\bf Theorem}[section]

\newtheorem{definition}{\bf Definition}[section]
\def\BibTeX{{\rm B\kern-.05em{\sc i\kern-.025em b}\kern-.08em
    T\kern-.1667em\lower.7ex\hbox{E}\kern-.125emX}}

\begin{document}

\title{Philosophy-Guided Mathematical Formalism for Complex Systems Modelling\\
\thanks{Corresponding author: Patrik Christen (patrik.christen@fhnw.ch).}
}

\author{\IEEEauthorblockN{Patrik Christen}
\IEEEauthorblockA{\textit{Institute for Information Systems} \\
\textit{FHNW}\\
Olten, Switzerland \\
patrik.christen@fhnw.ch}
\and
\IEEEauthorblockN{Olivier Del Fabbro}
\IEEEauthorblockA{\textit{Chair for Philosophy} \\
\textit{ETH Zurich}\\
Zurich, Switzerland \\
olivier.delfabbro@phil.gess.ethz.ch}
}

\maketitle

\begin{abstract}
We recently presented the so-called allagmatic method, which includes a system metamodel providing a framework for describing, modelling, simulating, and interpreting complex systems. Its development and programming was guided by philosophy, especially by Gilbert Simondon's philosophy of individuation, Alfred North Whitehead’s philosophy of organism, and concepts from cybernetics. Here, a mathematical formalism is presented to better describe and define the system metamodel of the allagmatic method, thereby further generalising it and extending its reach to a more formal treatment and allowing more theoretical studies. By using the formalism, an example for such a further study is provided with mathematical definitions and proofs for model creation and equivalence of cellular automata and artificial neural networks.
\end{abstract}


\section{Introduction}
Arguably every scientific discipline makes use of computer models, especially for the study of complex systems. Computational chemistry is a prime example since investigating chemical systems often involves phenomena at multiple length scales, which requires the combination of classical molecular mechanics and quantum mechanics. The success of computer modelling in that field is underlined by the Nobel Prize in Chemistry in 2013 that was awarded to work on a multiscale model for chemical systems \cite{RN203}. Another example is computational biology where, for example, numerical methods were combined with image analysis to show that bone cells are regulated by local mechanical loading in living humans \cite{RN29}. With the advent of agent-based modelling, also the social sciences started to make use of computer modelling \cite{RN215,Ansell.2016}. It seems to be a promising way to model complex human systems such as the economy \cite{RN212,Arthur.2014}. It is evident that computer modelling provides the sciences a powerful tool for enhanced and deeper studies of a wide range of systems, especially if they are complex.

Mathematical models form the theoretical basis for computer modelling. Classically since Newton and Leibniz, differential equations are usually used to describe the behaviour of a system at the macroscopic or global level \cite{RN238}. They allow predictions based on initial and boundary conditions, and many mathematical methods have been developed to study, solve, and approximate these equations. This analytical and global approach to mathematical modelling of systems has been and still is widely and successfully used in physics because interactions between individual components or particles of the modelled physical systems are usually homogeneous, constant in time, not specific to local elements, and linear \cite{RN238}. This is very different in complex systems, which are composed of a large number of elements characterised by states. These elements interact with one another in a spatially inhomogeneous, time-varying, element-specific, and often non-linear manner thereby forming a network structure \cite{RN238}. The dynamics of complex systems is therefore efficiently described by a list of rules updating element states at the local or microscopic level -- an algorithmic local approach used for mathematical modelling of systems \cite{RN238,RN207,RN288,RN427,RN358}. This means that the global behaviour of the system emerges from local interactions and is not explicitly described such as it is usually done with differential equations. There are certain mathematical models that describe the behaviour of local elements and their interactions, thus inherently taking into account that complex systems are modelled. E.g., cellular automata are known to model a variety of different systems including complex systems through simple local rules or transition functions \cite{RN25,RN55}. Agent-based models are a similar example of mathematical models describing systems algorithmically and locally. They are successfully used in the social sciences \cite{RN215,Ansell.2016}. A rather different example are artificial neural networks \cite{RN97,RN308}. They are different in the sense that they are used to uncover patterns in data and not to model the dynamics of systems unless the modelled system is a brain. Nevertheless, they consist of elements interacting with each other and the simulation result emerges from these local interactions.

However, mathematical formalisms and methods for studying complex systems algorithmically and locally are still scarce in comparison to analytical and global approaches. To investigate and better understand systems algorithmically and locally, in particular complex systems, we recently proposed the so-called allagmatic method to create computer models of systems \cite{RN114,RN210}. Allagmatic refers to exchange, alternation, or change of structures and operations \cite{Chateau.2008,DelFabbro.2021}. The method refers to changes and transformations a system is undergoing over time \cite{RN210,DelFabbro.2021,RN346} and furthermore borrows concepts from cybernetics \cite{DelFabbro.2021,RN59,RN71} and philosophy, particularly from the philosophy of individuation of the French philosopher Gilbert  Simondon \cite{DelFabbro.2021,Simondon.2020,RN125,delfabbro} and Alfred North Whitehead \cite{Whitehead.1929}. Simondon carefully and extensively studied technical objects such as engines and vacuum tubes, but also natural processes such as crystallisation and wave-particle dualism. He created a metaphysical and general terminology to describe these technical objects and natural processes.

In previous studies \cite{RN114,RN210}, these abstract concepts were implemented into the allagmatic method in the form of a system metamodel where technical and natural objects were looked at as systems. Rather than describing the global behaviour, the individual elements interacting with each other at a local level are described from which the global behaviour emerges. The global behaviour is thus implicitly defined. A system, according to the previous studies \cite{RN114,RN210}, consists of at least one structure describing the spatial domain and at least one operation describing the temporal domain of the system. Structure and operation together form a system in a so-called virtual regime providing an abstract description of a model of a system. Adding concrete parameters leads to the formation of a metastable system in a metastable regime providing a concrete model of the system. Finally, by adding further concrete parameters to the system and executing the model, an actual system in the actual regime is formed. 

We successfully used the system metamodel of the allagmatic method to create cellular automata and artificial neural networks \cite{RN210}, even via automatic programming \cite{RN114}. It therefore provides the basis for at least these two very different mathematical models and their concrete implementation. However, a rigorous mathematical formalism of the system metamodel is still missing. This would not only give a better description and definition of the system metamodel, it would furthermore allow generalisation and extension of its reach to a more formal treatment and more theoretical studies.

Note that the allagmatic method is generally useful for implementing hard to pin down concepts and exploring them and their relations to each other. In the present study, a detailed and rigorous mathematical formalism of the system metamodel of the allagmatic method is presented providing a formal basis for mathematical and theoretical study. Two such further studies are given: The creation of concrete mathematical models from the system metamodel and the conditions for equivalence of two models are mathematically defined and proved.

\section{The Allagmatic Method}

The system metamodel of the allagmatic method has been proposed to consist of model building blocks that correspond to Simondon's description of a system where at least one structure and one operation are combined together to form the system \cite{RN114,Simondon.2020,RN210}. Using notation of the previous studies, structure is represented by the local elements defined as a $p$-tuple $e=(e_1,e_2,e_3,\dots,e_{p})$ of $p$ \textit{entities} and their connection to neighbouring entities defined as a $q$-tuple \textit{milieu} $m_i$ of $q$ neighbours of the $i$-th entity $e_i$. Operation is represented by an \textit{update function} $\phi$ that updates the states of the entities over time based on the states of the neighbouring entities where $S$ is the set of possible states, e.g. $S=\{0,1\}$. $\phi$ has been defined as a mapping $\phi : S^{q+1} \rightarrow S$.

The model building blocks \textit{entity}, \textit{milieu}, and \textit{update function} were then implemented in C++ program code making use of object-oriented and generic programming \cite{RN210}. To describe an individual entity, the class \texttt{Entity} is implemented with the attribute \texttt{state} having a generic type. This is achieved through template meta-programming \cite{RN189}. To describe the system as a whole, the class \texttt{systemModel} is implemented with two attributes \texttt{systemEntities} and \texttt{milieus}. Both attributes are C++ vectors and have therefore a generic type and dynamic size. The milieu is represented by an adjacency matrix. The update function is implemented as a method called \texttt{updateFunction()} in classes representing specific systems or models. This implementation allows the definition of an abstract system in the virtual regime having no data types and sizes assigned yet and more concrete systems in the metastable regime by defining data types and sizes through concrete parameters.

A previous study showed that it is possible to use the allagmatic method to create simple cellular automata and artificial neural networks \cite{RN210}. In both cases, the models are given an input and asked to create a particular output. To achieve this, cellular automata with different local rules in the update function are created and evaluated in an evolutionary computation until the output is adequate. Similarly, artificial neural networks are created and learned until the output is acceptable. Both models successfully create the wanted output with an acceptable accuracy although it has to be stated that only simple models have been created so far.

In another experiment \cite{RN114}, the allagmatic method was used for automatic programming of mathematical models, i.e. cellular automata and artificial neural networks. This study made use of the general model building blocks combined to automatically create and implement models. The same experiments as in \cite{RN210} were performed but in an automatic manner forcing the program to create cellular automata and artificial neural networks. It is implemented in such a way that a string of program code is created based on variables, model building blocks, and specific update functions defined as string variables. The program code is then written to a file and compiled and executed using the \texttt{system} method in C++.

\section{Mathematical Formalism}

The allagmatic method has been introduced, described, its system metamodel implemented, and applied previously \cite{RN114,RN210}. However, some parts are not rigorously formulated yet, which would be important for the further studies of the allagmatic method. Especially the formalism of the system as a whole, the structure of the milieus and rules, and the evolutionary computation and learning methods still need to be defined mathematically, which is the purpose of this section.

\subsection{Notation}

In the following, \textit{tuples} are denoted with upper case letters in calligraphic font style and their elements indicated with corresponding lower case letters with hat accents written within parentheses, e.g. $a$-tuple $\mathcal{A}=(\hat{a}_1,\hat{a}_2,\hat{a}_3,\dots,\hat{a}_a)$, where the number of elements is indicated with corresponding lower case letters, e.g. $a$. Tuples in other tuples are denoted with upper case letters in calligraphic font style and with hat accents, e.g. $\hat{\mathcal{A}}$. \textit{Sets} are denoted with upper case letters and their elements denoted with corresponding lower case letters with tilde accents written within curly brackets, e.g. $B=\{ \tilde{b}_1,\tilde{b}_2,\tilde{b}_3,\dots,\tilde{b}_b \}$. \textit{Vectors} and \textit{matrices} are denoted with bold lower and upper case letters, respectively, e.g. $\mathbf{c}$ and $\mathbf{C}$. \textit{Functions} are denoted with lower case greek letters, e.g. $\alpha$. Furthermore, $\mathbb{N}$ refers to the set of positive integers $\{1,2,3,\dots\}$.

\subsection{Definitions}

If one wants to build a mathematical model to investigate a particular system, be it technical, social, or natural, the allagmatic method might be used. It allows the creation of a model of a system according to a system metamodel based on concepts borrowed from cybernetics \cite{RN59,RN71} and especially the philosophy of Simondon \cite{Simondon.2020,RN125,delfabbro}. Both suggest a system view of the studied objects and Simondon defines a system consisting of at least one structure and at least one operation leading to the following definition of a model of a system capturing the system as a whole:

\begin{definition}
	Let $\mathcal{SM}$ denote a model of a system, $S$ its set of structures or spatial domains, and $O$ its set of operations or temporal domains. Then $\mathcal{SM}$ is a $(s+o)$-tuple consisting of $s$ structures and $o$ operations, where $s, o\geq 1$ and therefore:
	\begin{equation}
		\mathcal{SM} \coloneqq (\hat{s}_1,\hat{s}_2,\hat{s}_3,\dots,\hat{s}_s,\hat{o}_1,\hat{o}_2,\hat{o}_3,\dots,\hat{o}_o),
	\end{equation}
	where $\hat{s}_i\in S$ and $\hat{o}_j\in O$.
\end{definition}

A tuple is used to describe the system as a whole similar to formal automata descriptions in discrete mathematics \cite{RN221}. Structures $S$ and operations $O$ can furthermore be defined more precisely and in the form of sets. According to the systems view, there are certain local elements that interact with each other. These elements or entities and also their interaction require a certain structure. As in our previous studies \cite{RN114,RN210}, all entities of a system can be defined with an entities $e$-tuple $\mathcal{E}=(\hat{e}_1,\hat{e}_2,\hat{e}_3,\dots,\hat{e}_{e})$, where $\hat{e}_i\in Q$ with $Q$ being the set of $k$ possible states. The milieus have been previously defined by an adjacency matrix $\mathbf{M}$ \cite{RN114,RN210}. If the system to be modelled is very large, $\mathbf{M}$ gets large as well, which can lead to performance issues, i.e. $\mathbf{M}$ not fitting into computer memory. The milieus $e$-tuple $\mathcal{M}=(\hat{\mathcal{M}}_1,\hat{\mathcal{M}}_2,\hat{\mathcal{M}}_3,\dots,\hat{\mathcal{M}}_{e})$ is therefore defined here as an adjacency list, where $\hat{\mathcal{M}}_{i}=(\hat{m}_1,\hat{m}_2,\hat{m}_3,\dots,\hat{m}_m)$ is the milieu of the $i$-th entity $\hat{e}_i$ of $\mathcal{E}$ consisting of $m$ neighbours of $\hat{e}_i$. Structures for update rules $\mathcal{U}$ and adaptation rules $\mathcal{A}$ are in this study introduced and defined as an $u$-tuple $\mathcal{U}$ and an $a$-tuple $\mathcal{A}$, respectively (see \cite{DelFabbro.2022} for more details on adaptation and control). Additionally, there is an adaptation end $p$-tuple $\mathcal{P}$ to capture the goal, target, or end system state of the adaptation process. $\mathcal{U}$ is related to the update function $\phi$ whereas $\mathcal{A}$ and $\mathcal{P}$ are newly introduced here and related to the adaptation function $\psi$. They are described and defined in more detail with respect to the definition of operation in the next paragraph. Some models might require to define further structures $\tilde{s}_i$ such as helper or counter variables. Taking this together, the set of structures $S$ can be defined as follows:

\begin{definition}
	The set of structures $S$ consists of the entities $e$-tuple $\mathcal{E}=(\hat{e}_1,\hat{e}_2,\hat{e}_3,\dots,\hat{e}_{e})$, where $\hat{e}_i$ is in the set $Q$ of $k$ possible states, the milieus $e$-tuple $\mathcal{M}=(\hat{\mathcal{M}}_1,\hat{\mathcal{M}}_2,\hat{\mathcal{M}}_3,\dots,\hat{\mathcal{M}}_{e})$, where $\hat{\mathcal{M}}_{i}=(\hat{m}_1,\hat{m}_2,\hat{m}_3,\dots,\hat{m}_m)$ is the milieu of the $i$-th entity $\hat{e}_i$ of $\mathcal{E}$ consisting of $m$ neighbours of $\hat{e}_i$, the update rules $u$-tuple $\mathcal{U}$, the adaptation rules $a$-tuple $\mathcal{A}$, the adaptation end $p$-tuple $\mathcal{P}$, and possibly further structures $\tilde{s}_i$, leading to the following definition:
	\begin{equation}
		S \coloneqq \{\mathcal{E}, Q, \mathcal{M}, \mathcal{U}, \mathcal{A}, \mathcal{P}, \dots, \tilde{s}_s \}.
	\end{equation}
\end{definition}

According to the recently introduced system metamodel \cite{RN114,RN210}, a model $\mathcal{SM}$ consists of at least one operation that operates on $\mathcal{E}$. More precisely, this is the update function $\phi : Q^{m+1} \rightarrow Q$ that updates the states of all entities $\mathcal{E}$ over a total number of $t$ discrete time steps. Its current time step is denoted with $\bar{t}$ and thus the update function's inputs are the states of entity $\hat{e}_i^{(\bar{t})}$ and its neighbours $\hat{\mathcal{M}}_i^{(\bar{t})}$ at time step $\bar{t}$ and the output is the new state of $\hat{e}_i^{(\bar{t}+1)}$ at time step $\bar{t}+1$, i.e. $\phi(\hat{e}_i,\hat{\mathcal{M}}_i,\bar{t})$. While we described and defined the update function earlier \cite{RN114,RN210}, the rules according to which the update function and the evolutionary and learning methods operate still need to be defined and described. In this study, the rules or the logic for the update function $\phi$ are defined as an $u$-tuple $\mathcal{U}$. E.g. for cellular automata, update rules $\mathcal{U}$ can be described with a truth table, which could be stored in the structure $\mathcal{U}$. In all of the previous experiments with the allagmatic method, some kind of adaptation, optimisation, or learning was used. In the case of cellular automata, this is an evolutionary computation and in the case of artificial neural networks it is a learning method. To account for that and to mathematically describe and define these kind of operations, the adaptation function $\psi$ is introduced in this study. This function also operates according to certain rules that need to be stored. For this, the adaptation rules $a$-tuple $\mathcal{A}$ is defined. If adaption is achieved through a mathematical function, the structure $\mathcal{A}$ might not be required, as the adaptation rules could be implemented implicitly in the function without requiring any additional data type and structure. Depending on the application, $\mathcal{SM}$ might not be capable of adapting and as such would not require an adaptation function $\psi$ and its adaptation rules $\mathcal{A}$. From this, the following definition for the set of operations $O$ can be formulated:

\begin{definition}
	The set of operations $O$ consists of, at least, an update function $\phi(\hat{e}_i,\hat{\mathcal{M}}_i,\bar{t},\mathcal{U})$, and optionally an adaptation function $\psi(\bar{g},\mathcal{A},\mathcal{P},l)$ as well as possible further operations $\tilde{o}_j$, leading to the following definition:
	\begin{equation}
		O \coloneqq \{ \phi(\hat{e}_i,\hat{\mathcal{M}}_i,\bar{t},\mathcal{U}),\psi(\bar{g},\mathcal{A},\mathcal{P},l),\dots, \tilde{o}_o \},
	\end{equation}
	where $\bar{t}$ is the current time step, $\bar{g}$ the current adaptation iteration, and $l$ the loss tolerance.
\end{definition}

Based on the definitions of the set of structures $S$ and the set of operations $O$, the definition of the system model $\mathcal{SM}$ can be revised as follows:

\begin{definition}
	Let $\mathcal{SM}$ denote a model of a system, $S$ its set of structures or spatial domains capturing entities in $\mathcal{E}$ and their possible states in $Q$ and local neighbourhood or milieu in $\mathcal{M}$ and structures related to update function in $\mathcal{U}$ and to the adaptation function in $\mathcal{A}$ and $\mathcal{P}$, and $O$ its set of operations or temporal domains capturing the dynamic state transitions of the entities $\mathcal{E}$ with the function $\phi$ and possible adaptations or evolutions of the system with the function $\psi$. Then $\mathcal{SM}$ is a $(s+o)$-tuple consisting of $s$ structures and $o$ operations, where $s,o\geq 1$ and therefore:
	\begin{equation}
		\mathcal{SM} \coloneqq (\mathcal{E},Q,\mathcal{M},\mathcal{U},\mathcal{A},\mathcal{P},\dots,\hat{s}_s,\phi,\psi,\dots,\hat{o}_o),
	\end{equation}
where $\hat{s}_i\in S$ and $\hat{o}_j\in O$.
\end{definition}

\subsection{Concrete Parameters}

All structural and operational definitions of the allagmatic method are provided in the previous section. They are an abstract description of a system model and therefore represent a virtual system in the virtual regime. Concrete parameters are then fed into the virtual system to create a more concrete model, the metastable system in the metastable regime. These concrete parameters completely determine the modelled system $\mathcal{SM}$. Specifically, with respect to structures $S$, they are the entities $e$-tuple $\mathcal{E}$ including the initial states $\mathcal{E}_0$ and therefore also $Q$, the milieus $e$-tuple $\mathcal{M}$ including its values, structures and specifications for the update rules $u$-tuple $\mathcal{U}$ and, if considered, the adaptation rules $a$-tuple $\mathcal{A}$ and adaption end $p$-tuple $\mathcal{P}$. With respect to operations $O$, they are the specifications of the update function $\phi$ and, if considered, the adaptation function $\psi$. Besides these concrete parameters, there are also simple numbers indicating the number of structures or operations that can also be considered as concrete parameters. They are the number of structures $s$, operations $o$, entities $e$, possible entity states $k$, neighbours $m$, update rules $u$, adaptation rules $a$, discrete time steps $t$, and adaptation iterations $g$. With the exception of $t$ and $g$, these concrete parameters can be extracted from other concrete parameters. The complete list of concrete parameters is summarised in Table ~\ref{tab:concreteParameters}.

\begin{table}[!h]
\centering
 \caption{Concrete parameters of the allagmatic method.}
 \label{tab:concreteParameters}
\begin{tabular}{m{0.005\linewidth}m{0.38\linewidth}m{0.45\linewidth}}
   \hline
   $\mathcal{E}$ 	&	entities				&	$\mathcal{E}$ is an $e$-tuple where $\hat{e}_i\in Q$								\\
   $Q$	&	possible entity states		&	$Q$ is a set where $|Q|=k$								\\
   $\mathcal{M}$	&	milieus				&	$\mathcal{M}$	is an $e$-tuple where $\hat{\mathcal{M}}_i$ is a tuple with $m$ elements $\in \mathcal{E}$							\\
   $\mathcal{U}$	&	update rules			&	$\mathcal{U}$ is an $u$-tuple								\\
   $\mathcal{A}$	&	adaptation rules		&	$\mathcal{A}$ is an $a$-tuple								\\
   $\mathcal{P}$	&	adaptation end		&	$\mathcal{P}$ is an $p$-tuple								\\
   $\phi$	&	update function			&	$\phi : Q^{m+1} \rightarrow Q$									\\
   $\psi$	&	adaptation function		&	e.g. an evolutionary computation								\\
   \hline
   $s$	&	number of structures		& 	$\{s\in\mathbb{N}\mid s\geq1\}$		\\
   $o$	&	number of operations	& 	$\{o\in\mathbb{N}\mid o\geq1\}$		\\
   $e$	&	number of entities		& 	$e\in\mathbb{N}$		\\
   $k$	&	number of possible entity states	& 	$k\in\mathbb{N}$		\\
   $m$	&	number of neighbours	& 	$m\in\mathbb{N}$		\\
   $u$	&	number of update rules	& 	$u\in\mathbb{N}$		\\
   $a$	&	number of adaptation rules	& 	$a\in\mathbb{N}$		\\
   $p$	&	number of adaptation ends	& 	$p\in\mathbb{N}$		\\
   $l$		&	adaption loss tolerance		& 	$l\in\mathbb{R}$		\\
   $t$		&	number of discrete time steps	& 	$t\in\mathbb{N}$		\\
   $\bar{t}$		&	current time step	& 	$\bar{t}\in\mathbb{N}$		\\
   $g$	&	number of adaptation iterations	& 	$g\in\mathbb{N}$		\\
   $\bar{g}$	&	current adaptation iteration	& 	$\bar{g}\in\mathbb{N}$		\\
   \hline
 \end{tabular}
 \vspace*{-4pt}
 \end{table}

\section{Model Creation}

Our two recent studies \cite{RN114,RN210} have already showed that the system metamodel of the allagmatic method can be used to create concrete mathematical models, specifically cellular automata and artificial neural networks. This was achieved by computational means and thus by programming of the concrete models according to the system metamodel. A more formal and precise account of this model creation is given here based on the definitions from the previous section. The mathematical description of the system metamodel is first used to create cellular automata and second to create artificial neural networks. It can thus be regarded as a first case of the presented formalism to a more formal and theoretical study of the allagmatic method.

\subsection{Cellular Automata}
Although simple, cellular automata were successfully used to explore complex problems from their discovery to the present day. John von Neumann discovered cellular automata under the influence of Stanislaw Ulam \cite{RN204} and succeeded in creating a system capable of self-replication \cite{RN205,Burks.1970}. With his \textit{Game of Life} \cite{RN206}, John H. Conway not only created the arguably most popular automaton, he also provided one of the simplest models of computation to be universal \cite{sep-cellular-automata}. An extensive study of simple one-dimensional cellular automata followed by Stephen Wolfram, where he provided qualitative taxonomy of cellular automata behaviour \cite{RN25}, especially complex behaviour \cite{RN55}. He also conjectured and provided a sketch of a proof \cite{RN25} that one particular transition function called rule 110 is capable of universal computation, which was indeed proved later \cite{RN96}. Besides using cellular automata as models for computation, they have also been suggested to be an alternative for differential equations in the modelling of physical systems \cite{RN207} and with that open up a wide range of applications.

Different types of cellular automata have been proposed and described in many different ways depending on the field and author \cite{RN16,RN35}. To provide a general and broadly applicable definition of cellular automata, the respective entry in the \textit{Stanford Encyclopedia of Philosophy} \cite{sep-cellular-automata} is used in this paper as a description and definition of cellular automata. Cellular automata are discrete, abstract computational systems. They are spatially and temporally discrete and composed of simple units, the atoms or cells. These cells instantiate one of a finite set of states evolving over discrete time steps according to state update functions or dynamical transition rules that take into account the local neighbourhood of each cell.

Although many systems have been defined as cellular automaton, there are four elements that describe virtually every cellular automaton. First, there is a discrete lattice of $c$ cells denoted with $\mathcal{C}$. It is described as an ordered list and thus $c$-tuple $\mathcal{C}$. Second, each of these cells is in a state of $k$ possible states $K$ at each discrete time step $\bar{t}$, thus $\hat{c}_i \in K$. Third, only local interactions and therefore the states of $n$ neighbouring cells $\hat{\mathcal{N}}_i=(\hat{n}_1,\hat{n}_2,\hat{n}_3,\dots,\hat{n}_n)$ of $\hat{c}_i$ and possibly the cell itself determine the behaviour of an individual cell. Note that the $c$-tuple $\mathcal{N}$ contains the neighbouring cells of every cell. And fourth, each cell is updated at each discrete time step $\bar{t}$ according to a deterministic transition function $\delta : K^{n} \rightarrow K$ or $\delta : K^{n+1} \rightarrow K$ if besides the neighbours also the state of the cell itself is taken into account. A cellular automaton can thus be defined as follows:

\begin{definition}
	A cellular automaton $\mathcal{CA}$ is defined as a $4$-tuple consisting of a $c$-tuple $\mathcal{C}$ with $c$ cells or elements, a set of discrete states $K$, where $\hat{c}_i \in K$, a $c$-tuple $\mathcal{N}$ with local interactions for each $\hat{c}_i$ of $\mathcal{C}$, and a deterministic transition function $\delta : K^{n+1} \rightarrow K$, where $n$ indicates the number of neighbours:
	\begin{equation}
		\mathcal{CA} \coloneqq ( \mathcal{C},K,\mathcal{N},\delta ).
	\end{equation}
\end{definition}

With these definitions, it is now possible to show that cellular automata are contained in the system model:

\begin{theorem}
A cellular automaton $\mathcal{CA}$ can be created from the system model $\mathcal{SM}$, thus $\mathcal{CA}$ is a special case of $\mathcal{SM}$ under the conditions of no adaptation function $\psi$ and further structures $\hat{s}_i$ and operations $\hat{o}_j$, $\mathcal{U}$ is implicitly specified as part of $\delta$, and an equivalent mapping implementation of $\phi$ and $\delta$.\end{theorem}

\begin{proof}
Since $\mathcal{CA} \coloneqq ( \mathcal{C},K,\mathcal{N},\delta )$, a system model $\mathcal{SM}^{CA}$ describing a $\mathcal{CA}$ needs no adaptation function $\psi$, further structures $\hat{s}_i$ and operations $\hat{o}_j$, and $\mathcal{U}$ is implicitly specified as part of $\delta$, which leads to $\mathcal{SM}^{CA} = (\mathcal{E},Q,\mathcal{M},\phi)$. $\mathcal{SM}^{CA} = \mathcal{CA} \iff \mathcal{E}=\mathcal{C}, Q=K,\mathcal{M}=\mathcal{N},\phi=\delta$. $\mathcal{E}=\mathcal{C} \iff \forall i : 1\leq i \leq x : \hat{e}_i = \hat{c}_i$, where $i$ denotes the $i$-th element and $x$ the number of elements. This is true since $x=e=c$ and $Q=K$. The latter is shown with $Q=K \iff (\forall y : y \in Q \iff y \in K)$, where $y$ denotes an element. This is true since $|Q|=|K|$ and, noticing that $\mathcal{E}=\mathcal{C}$, $\hat{e}_i\in Q$ and $\hat{c}_i \in K$. $\mathcal{M}=\mathcal{N} \iff \forall i : 1\leq i \leq z : \hat{\mathcal{M}}_i = \hat{\mathcal{N}}_i$, where $i$ denotes the $i$-th element and $z$ the number of elements. This is true since $z=e=c$ and, noticing that $n$ is the same for all $\hat{c}_i$ in $\mathcal{CA}$, $|\hat{\mathcal{M}}_i|=m=|\hat{\mathcal{N}}_i|=n$. Elements of $\hat{\mathcal{M}}_i$ are in $\mathcal{E}$ and elements of $\hat{\mathcal{N}}_i$ are in $\mathcal{C}$. Since $\mathcal{E}=\mathcal{C}$, $\hat{\mathcal{M}}_i$ and $\hat{\mathcal{N}}_i$ have elements in the same set. $\phi:Q^{m+1}\rightarrow Q = \delta : K^{n+1} \rightarrow K \iff Q=K, \forall b \in Q:\phi(b)=\delta(b)$. This is true since $Q=K$ and $m=n$ have already been shown and because the mapping can be implemented in $\mathcal{SM}^{CA}$ and $\mathcal{CA}$ such that $\forall b \in Q:\phi(b)=\delta(b)$. This proves the theorem and gives the following conditions where $\mathcal{CA}$ is a special case of $\mathcal{SM}$: no adaptation function $\psi$ and further structures $\hat{s}_i$ and operations $\hat{o}_j$, $\mathcal{U}$ is implicitly specified as part of $\delta$, and an equivalent mapping implementation of $\phi$ and $\delta$.
\end{proof}

\subsection{Artificial Neural Networks}
A variety of artificial neural networks have been presented and applied, especially with respect to their topology as nicely illustrated by The Neural Network Zoo \cite{ann_zoo}. However, many textbooks from university courses have been developed over the years, e.g. \textit{Artificial Intelligence: A Modern Approach} from Stuart J. Russel and Peter Norvig \cite{RN97} that will serve here for the description and definition of artificial neural networks. Artificial neural networks consist of a mathematical model of neurons that are connected together forming a network where a certain output is computed according to the given input into the network.

In general, artificial neural networks are composed of units or neurons $\mathcal{B}$ connected by directed links, where a link from unit $\hat{b}_i$ to unit $\hat{b}_j$ propagates the so-called activation $a_i$. The activation refers to the signal transmitted by a neuron, which also has a weight $w_{i,j}$, representing the strength of the signal between $\hat{b}_i$ and $\hat{b}_j$. Inside every neuron $\hat{b}_j$, the weighted sum of $r$ input activation signals $a_i$ from other neurons $\hat{b}_i$ is calculated with an input function $\beta{_j}=\sum_{i=1}^r{w_{i,j} a_i}$. Each neuron also consists of an activation function $\alpha(\beta{_j})=a_j$ calculating the activation signal $a_j$, the output of $\hat{b}_j$. $\alpha$ is typically either a threshold or logistic function. These units are usually arranged into multi-layer neural networks with input, middle or hidden, and output layers. Based on training data sets, the output of the network can be calculated and compared to the desired output, allowing to calculate an error. This error is backpropagated through the network layers and the weights $w$ are adapted according to a certain learning rule adapting the weights $w$ \cite{RN97}.

More formally, units $\mathcal{B}$ are described with a $b$-tuple, where $b$ is the number of units, their possible values with a set $V$, incoming activation units with a $b$-tuple $\mathcal{R}$ that consists, for each unit $\hat{b}_j$, an $r$-tuple $\hat{\mathcal{R}}_j$, activation with a function composition $\alpha\circ \beta{_j}$, and learning with a learning rule or function $\xi$ adapting the weights $w$. Therefore, the following definition of an artificial neural network can be given:

\begin{definition}
	An artificial neural network $\mathcal{ANN}$ is defined as a $5$-tuple consisting of a units $b$-tuple $\mathcal{B}$ with $b$ units, a set $V$ of possible values for $\hat{b}_j$ and thus $\hat{b}_j\in V$, a $b$-tuple $\mathcal{R}$ with incoming activation units for each $\hat{b}_j$, a function composition $\alpha\circ \beta{_j}:V^r\rightarrow V$ determining the activation signal, where $r$ is the number of incoming activation units, and a learning function $\xi:\mathbb{R}\rightarrow\mathbb{R}$ adapting $w$:
	\begin{equation}
		\mathcal{ANN} \coloneqq \left(\mathcal{B},V,\mathcal{R},\alpha\circ \beta{_j},\xi  \right).
	\end{equation}
\end{definition}

Given the system model $\mathcal{SM}$ as defined in the previous section, it should now be possible to arrive at $\mathcal{ANN}$ as defined above providing the mathematical creation of an artificial neural network from the system metamodel of the allagmatic method. The following theorem can thus be stated:

\begin{theorem}
An artificial neural network $\mathcal{ANN}$ can be created from the system model $\mathcal{SM}$, thus $\mathcal{ANN}$ is a special case of $\mathcal{SM}$ under the conditions of no further structures $\hat{s}_i$ and operations $\hat{o}_j$ as well as $\mathcal{U}$ and $\mathcal{A}$ are implicitly specified as part of $\alpha\circ \beta{_j}$ and $\xi$, respectively, and an equivalent mapping implementation of $\phi$ and $\alpha\circ \beta{_j}$ as well as $\psi$ and $\xi$.
\end{theorem}

\begin{proof}
$\mathcal{ANN} \coloneqq \left(\mathcal{B},V,\mathcal{R},\alpha\circ \beta{_j},\xi  \right)$ implies no further structures $\hat{s}_i$ and operations $\hat{o}_j$ as well as $\mathcal{U}$ and $\mathcal{A}$ are implicitly specified as part of $\alpha\circ \beta{_j}$ and $\xi$, respectively, which leads to $\mathcal{SM}^{ANN} = (\mathcal{E},Q,\mathcal{M},\phi,\psi)$. $\mathcal{SM}^{ANN} = \mathcal{ANN} \iff \mathcal{E}=\mathcal{B}, Q=V, \mathcal{M}=\mathcal{R}, \phi=\alpha\circ \beta{_j}, \psi=\xi$. $\mathcal{E}=\mathcal{B} \iff \forall i : 1\leq i \leq x : \hat{e}_i = \hat{b}{_i}$, where $i$ denotes the $i$-th element and $x$ the number of elements. This is true since $x=e=b$ and $Q=V$. The latter is shown with $Q=V \iff (\forall y : y \in Q \iff y \in V)$, where $y$ denotes an element. This is true since $|Q|=|V|$ and, noticing that $\mathcal{E}=\mathcal{B}$, $\hat{e}_i\in Q$ and $\hat{b}{_i} \in V$. $\mathcal{M}=\mathcal{R} \iff \forall i : 1\leq i \leq z : \hat{\mathcal{M}}_i = \hat{\mathcal{R}}_i$, where $i$ denotes the $i$-th element and $z$ the number of elements. This is true since $z=e=b$ and elements of $\hat{\mathcal{M}}_i$ are in $\mathcal{E}$ and elements of $\hat{\mathcal{R}}_i$ are in $\mathcal{B}$. Since $\mathcal{E}=\mathcal{B}$, $\hat{\mathcal{M}}_i$ and $\hat{\mathcal{R}}_i$ have elements in the same set. $\phi:Q^{m+1}\rightarrow Q = \alpha\circ \beta{_j}:V^r\rightarrow V \iff Q=V, \forall b \in Q:\phi(b)=(\alpha\circ \beta{_j})(b)$. This is true since $Q=V$ has been shown already and because the mapping can be implemented in $\mathcal{SM}^{ANN}$ and $\mathcal{ANN}$ such that $\forall b \in Q:\phi(b)=(\alpha\circ \beta{_j})(b)$. $\psi=\xi \iff \forall b:\psi(b)=\xi(b)$ can be implemented in $\mathcal{SM}^{ANN}$ and $\mathcal{ANN}$ that it is true as well. This proves the theorem and gives the conditions where $\mathcal{ANN}$ is a special case of $\mathcal{SM}$: No further structures $\hat{s}_i$ and operations $\hat{o}_j$ as well as $\mathcal{U}$ and $\mathcal{A}$ are implicitly specified as part of $\alpha\circ \beta{_j}$ and $\xi$, respectively, and an equivalent mapping implementation of $\phi$ and $\alpha\circ \beta{_j}$ as well as $\psi$ and $\xi$.
\end{proof}

\section{Model Equivalence}

Not only the creation of concrete mathematical models is possible with the system metamodel, also the comparison of such models. Such a comparison gives guidance of how different models can be applied to the same system or problem but in an equivalent manner. E.g. model parameters are mapped to the same structures and operations of the system being modelled. It focuses on figuring out the differences or missing parts between two models if applied to the same system or problem providing therefore the conditions for model equivalence. The comparison between models is achieved by first mapping the structures and operations of the two models to be compared to the system metamodel. This is the model creation proof presented above. From the model creation proofs, it then becomes immediately evident which structures and operations of the two models map to the same structures and operations of the system metamodel and which ones are missing. Matching structures and operations show which parts of the two models have to have the same type or input-output mapping, respectively, and the missing structures and operations show the conditions for model equivalence. More precisely, we can define model equivalence in the context of the allagmatic method as follows:

\begin{definition}
	Two models are equivalent in the context of the allagmatic method if they can be created with the same structures $\mathcal{S}$ and operations $\mathcal{O}$ from the system metamodel $\mathcal{SM}$. Thereby, structures need to be of the same type and operations need to have the same input-output mapping. Non-matching or missing structures and operations indicate conditions required for model equivalence, which might be called conditional equivalence.
\end{definition}

In the following, a first example for a model comparison is given by showing under which conditions a cellular automaton is structurally and operationally equivalent to an artificial neural network. Based on the presented formalism of the system metamodel and the model creation proofs of cellular automata and artificial neural networks, the theorem below can thus be stated:

\begin{theorem}
A cellular automaton $\mathcal{CA}$ is conditionally equivalent to an artificial neural network $\mathcal{ANN}$ if an equivalent mapping of $\delta$ and $\alpha\circ \beta{_j}$ is implemented and where the missing $\xi$ in $\mathcal{CA}$ forms the condition for conditional model equivalence.
\end{theorem}

\begin{proof}
We know that $\mathcal{SM}^{CA} = \mathcal{CA} \iff \mathcal{E}=\mathcal{C}, Q=K,\mathcal{M}=\mathcal{N},\phi=\delta$ and $\mathcal{SM}^{ANN} = \mathcal{ANN} \iff \mathcal{E}=\mathcal{B}, Q=V, \mathcal{M}=\mathcal{R}, \phi=\alpha\circ \beta{_j}, \psi=\xi$. It follows that $\mathcal{CA} = \mathcal{ANN} \iff \mathcal{C}=\mathcal{B}, K=V, \mathcal{N}=\mathcal{R}, \delta=\alpha\circ \beta{_j}$ and that $\xi$ of $\mathcal{ANN}$ is missing in $\mathcal{CA}$. $\mathcal{C}=\mathcal{B} \iff \forall i : 1\leq i \leq x : \hat{c}_i = \hat{b}{_i}$, where $i$ denotes the $i$-th element and $x$ the number of elements. This is true since $x=c=b$ and $K=V$. The latter is shown with $K=V \iff (\forall y : y \in K \iff y \in V)$, where $y$ denotes an element. This is true since $|K|=|V|$ and, noticing that $\mathcal{C}=\mathcal{B}$, $\hat{c}_i\in K$ and $\hat{b}{_i} \in V$. $\mathcal{N}=\mathcal{R} \iff \forall i : 1\leq i \leq z : \hat{\mathcal{N}}_i = \hat{\mathcal{R}}_i$, where $i$ denotes the $i$-th element and $z$ the number of elements. This is true since $z=c=b$ and elements of $\hat{\mathcal{N}}_i$ are in $\mathcal{C}$ and elements of $\hat{\mathcal{R}}_i$ are in $\mathcal{B}$. Since $\mathcal{C}=\mathcal{B}$, $\hat{\mathcal{N}}_i$ and $\hat{\mathcal{R}}_i$ have elements in the same set. $\delta:K^{n+1}\rightarrow K = \alpha\circ \beta{_j}:V^r\rightarrow V \iff K=V, \forall b \in K:\delta(b)=(\alpha\circ \beta{_j})(b)$. This is true since $K=V$ has been shown already and because the mapping can be implemented in $\mathcal{CA}$ and $\mathcal{ANN}$ such that $\forall b \in K:\delta(b)=(\alpha\circ \beta{_j})(b)$. The missing $\xi$ in $\mathcal{CA}$ could be created with $\psi$ from $\mathcal{SM}$ and implemented in such a way that $\psi=\xi \iff \forall b:\psi(b)=\xi(b)$. However, here it forms the condition for conditional model equivalence. This proves the theorem and gives the conditions where $\mathcal{CA}$ is conditionally equivalent to $\mathcal{ANN}$: an equivalent mapping implementation of $\delta$ and $\alpha\circ \beta{_j}$ and the missing $\xi$ in $\mathcal{CA}$ forming the condition for conditional model equivalence.
\end{proof}

\section{Conclusion and Outlook}

We have recently proposed a system metamodel that seems especially promising to capture complex behaviour due to its algorithmic and local nature \cite{RN114,RN210}. We have already showed that the system metamodel can be used to create concrete mathematical models guiding their implementation. This was done by computational means implementing cellular automata and artificial neural networks as a running computer model from the system metamodel. In the present paper, a rigorous mathematical formalism of the system metamodel is presented to better describe, define, and generalise the system metamodel. The formalism is used to prove model creation and equivalence of cellular automata and artificial neural networks.

The definitions from our previous work \cite{RN114,RN210} are extended in this study, especially with respect to the system as a whole, the structure of the milieus and rules, and the evolutionary computation and learning methods. The system as a whole and thus model of a system is defined as a tuple $\mathcal{SM}$ consisting of structures $\hat{s}_i\in S$ and operations $\hat{o}_j\in O$, more concretely $\mathcal{SM} \coloneqq (\mathcal{E},Q,\mathcal{M},\mathcal{U},\mathcal{A},\dots,\hat{s}_s,\phi,\psi,\dots,\hat{o}_o)$. Where $\mathcal{E}$ is an $e$-tuple of $e$ entities, $Q$ is a set of $k$ possible entity states and thus $\hat{e}_i\in Q$, $\mathcal{M}$ is an $e$-tuple of $e$ milieus $\hat{\mathcal{M}}_i$ that are tuples consisting of the milieu or neighbours of the corresponding entities $\hat{e}_i$, $\mathcal{U}$ is a structure related to the update function $\phi$ and $\mathcal{A}$ related to the adaptation function $\psi$, and $\hat{s}_i$ and $\hat{o}_j$ are any further structures and operations, respectively. With this definition, the most fundamental elements of a system model are described but it is still possible to include further structures and operations. Since each of these fundamental structures and operations have their own role, the order matters and thus a mathematical tuple seems adequate for the formal description of the system as a whole. Furthermore, it is common to describe automata as a whole with tuples in discrete mathematics \cite{RN221}. In contrary, to describe structures and operations, it seems more useful to refer to either a structure or an operation and thus a set is more suitable for the description of structural and operational objects. For the milieu $\mathcal{M}$, two nested tuples are used, which is a flexible yet efficient way to describe the topology of connected entities. With respect to operations, there are at least two fundamental operations possibly occurring in a system. An operation that changes the states of entities, which has been already defined as $\phi$ in the previous studies \cite{RN114,RN210}. In addition, there are also operations changing certain parameters of the system. Such operations are newly introduced here and called adaptation functions denoted as $\psi$. Evolutionary computations with cellular automata or learning in artificial neural networks are two examples of adaptation functions. In the former, the rules of the update function $\phi$ are changed and in the latter the weights $w$. Structures $\mathcal{U}$ and $\mathcal{A}$ might not be explicitly stated since they are assumed to be part of $\phi$ and $\psi$, respectively. The extension of the allagmatic method with an adaptation function allows now to describe adaptive, evolutionary, and reconfigurable systems.

Based on this definition of $\mathcal{SM}$, the creation of cellular automata and artificial neural networks are proved. Cellular automata creation is proved under the conditions of no adaptation function $\psi$ and further structures $\hat{s}_i$ and operations $\hat{o}_j$, $\mathcal{U}$ is implicitly specified as part of $\delta$, and an equivalent mapping implementation of $\phi$ and $\delta$, where $\delta$ is the update function of local states in cellular automata and thus equivalent to $\phi$. Cellular automata are sometimes coupled with an evolutionary computation \cite{RN344} as this was also done in the previous studies with the allagmatic method \cite{RN114,RN210}. To model the evolutionary computation, the adaptation function $\psi$ would also be specified in addition to $\phi$. Artificial neural networks creation is proved under the conditions of no further structures $\hat{s}_i$ and operations $\hat{o}_j$ as well as $\mathcal{U}$ and $\mathcal{A}$ are implicitly specified as part of $\alpha\circ \beta{_j}$ and $\xi$, respectively, and an equivalent mapping implementation of $\phi$ and $\alpha\circ \beta{_j}$ as well as $\psi$ and $\xi$. For both, cellular automata and artificial neural networks, concrete values of $\hat{e}_i$ and mappings are not directly proved since the creation of a model from a metamodel is proved and not the simulation of the model. However, they involve equivalent sets. The model creation proofs of cellular automata and artificial neural networks are then used in the model equivalence proof between the two models. The conditions in that proof are an equivalent mapping implementation of $\delta$ and $\alpha\circ \beta{_j}$ and the missing $\xi$ in $\mathcal{CA}$ forms the conditions for conditional model equivalence.

The presented definitions in this study are precise and rigorous descriptions of the system metamodel as a mathematical object. Whitehead was right by saying that philosophy has “[...] to insist on the scrutiny of the ultimate ideas, and on the retention of the whole of the evidence in shaping our cosmological scheme.” \cite{Whitehead.1967} But “mathematics is the science of the most complete abstractions to which the human mind can attain.” \cite{Whitehead.1967} In other words: Mathematics makes philosophical conceptions more precise and accurate, because it explores the interconnections between ultimate abstractions.

Some missing or vaguely described parts of the system metamodel are now included into the mathematical formalism by the presented definitions. The two model creation proofs show that the system metamodel of the allagmatic method can be used to create at least two very different mathematical models, i.e. cellular automata and artificial neural networks. The model equivalence proof shows that the conditions under which these two models are equivalent can be found via the system metamodel. As clearly defined mathematical objects, the definitions now allow such mathematical treatment, which builds the formal basis for the presented proofs. They also further generalise the system metamodel extending its reach to a more formal treatment and allowing more theoretical studies.

\section*{Acknowledgment}
%
We thank Dr. Marcel Wirz for helping with mathematical notation and formalism.

\bibliographystyle{ieeetr}
\bibliography{CS_Formalism_SMC}

\begin{thebibliography}{10}

\bibitem{RN203}
W.~Thiel and G.~Hummer, ``Methods for computational chemistry,'' {\em Nature},
  vol.~504, no.~7478, pp.~96--97, 2013.

\bibitem{RN29}
P.~Christen, K.~Ito, R.~Ellouz, S.~Boutroy, E.~Sornay-Rendu, R.~D. Chapurlat,
  and B.~van Rietbergen, ``Bone remodelling in humans is load-driven but not
  lazy,'' {\em Nature Communications}, vol.~5, 2014.

\bibitem{RN215}
E.~Bonabeau, ``Agent-based modeling: Methods and techniques for simulating
  human systems,'' {\em Proceedings of the National Academy of Sciences of the
  United States of America}, vol.~99, pp.~7280--7287, 2002.

\bibitem{Ansell.2016}
C.~Ansell and R.~Geyer, ``{‘Pragmatic complexity’ a new foundation for
  moving beyond ‘evidence-based policy making’?},'' {\em Policy Studies},
  pp.~1--19, 2016.

\bibitem{RN212}
J.~D. Farmer and D.~Foley, ``The economy needs agent-based modelling,'' {\em
  Nature}, vol.~460, no.~7256, pp.~685--686, 2009.

\bibitem{Arthur.2014}
W.~B. Arthur, {\em {Complexity and the Economy}}.
\newblock New York, NY: Oxford University Press, 2014.

\bibitem{RN238}
S.~Thurner, R.~Hanel, and P.~Klimek, {\em {Introduction to the Theory of
  Complex Systems}}.
\newblock New York, NY: Oxford University Press, 2018.

\bibitem{RN207}
T.~Toffoli, ``Cellular automata as an alternative to (rather than an
  approximation of) differential-equations in modeling physics,'' {\em Physica
  D}, vol.~10, no.~1-2, pp.~117--127, 1984.

\bibitem{RN288}
E.~Rauch, ``Discrete, amorphous physical models,'' {\em International Journal
  of Theoretical Physics}, vol.~42, no.~2, pp.~329--348, 2003.

\bibitem{RN427}
J.~H. Holland, {\em Emergence: From Chaos to Order}.
\newblock New York, NY: Oxford University Press, 1998.

\bibitem{RN358}
J.~H. Holland, {\em Signals and Boundaries: Building Blocks for Complex
  Adaptive Systems}.
\newblock Cambridge, MA: MIT Press, 2012.

\bibitem{RN25}
S.~Wolfram, {\em A New Kind of Science}.
\newblock Champaign, IL: Wolfram Media, 2002.

\bibitem{RN55}
S.~Wolfram, ``Cellular automata as models of complexity,'' {\em Nature},
  vol.~311, no.~5985, pp.~419--424, 1984.

\bibitem{RN97}
S.~J. Russel and P.~Norvig, {\em Artificial Intelligence: A modern Approach}.
\newblock Upper Saddle River, NJ: Prentice Hall, 2010.

\bibitem{RN308}
M.~P. Deisenroth, A.~A. Faisal, and C.~S. Ong, {\em Mathematics for Machine
  Learning}.
\newblock Cambridge and New York: Cambridge University Press, 2020.

\bibitem{RN114}
P.~Christen and O.~Del~Fabbro, ``Automatic programming of cellular automata and
  artificial neural networks guided by philosophy,'' in {\em New Trends in
  Business Information Systems and Technology} (R.~Dornberger, ed.), vol.~294
  of {\em Studies in Systems, Decision and Control}, pp.~131--146, Cham:
  Springer, 2021.
\newblock arXiv:1905.04232 [cs.AI].

\bibitem{RN210}
P.~Christen and O.~Del~Fabbro, ``{Cybernetical Concepts for Cellular Automaton
  and Artificial Neural Network Modelling and Implementation},'' in {\em 2019
  IEEE International Conference on Systems, Man and Cybernetics (IEEE SMC)},
  pp.~4124--4130, 2019.
\newblock arXiv:2001.02037 [cs.OH].

\bibitem{Chateau.2008}
J.-Y. Chateau, {\em Le vocabulaire de Simondon}.
\newblock Paris: Ellipses, 2008.

\bibitem{DelFabbro.2021}
O.~Del~Fabbro, {\em {Philosophieren mit Objekten: Gilbert Simondons prozessuale
  Individuationsontologie}}.
\newblock Frankfurt and New York: Campus Verlag, 2021.

\bibitem{RN346}
G.~Simondon, ``Allagmatique,'' in {\em L'individuation \`{a} la lumi\`{e}re des
  notions de forme et d'information}, pp.~529--536, Grenoble: Editions
  J\'{e}r\^{o}me Millon, 2013.

\bibitem{RN59}
W.~R. Ashby, {\em An Introduction to Cybernetics}.
\newblock London: Chapman \& Hall, 1956.

\bibitem{RN71}
N.~Wiener, {\em Cybernetics: Or Control and Communication in the Animal and the
  Machine}.
\newblock Cambridge, MA: MIT Press, 1961.

\bibitem{Simondon.2020}
G.~Simondon, {\em {Individuation in Light of Notions of Form and Information
  (T. Atkins, trans.)}}, vol.~I \& II.
\newblock Minneapolis and London: University Minnesota Press, 2020.

\bibitem{RN125}
G.~Simondon, {\em On the Mode of Existence of Technical Objects}.
\newblock Minneapolis: Univocal Publishing, 2017.

\bibitem{delfabbro}
O.~Del~Fabbro, ``Relationale {E}xistenzweisen von {M}aschinen,'' in {\em
  Mensch-Maschine-Interaktion} (K.~Liggieri and O.~M\"uller, eds.), pp.~63--70,
  Stuttgart: J.B. Metzler, 2019.

\bibitem{Whitehead.1929}
A.~N. Whitehead, {\em {Process and Reality: An Essay in Cosmology (D. R. Grifin
  and D. W. Sherburne, eds.)}}.
\newblock New York, NY: Free Press, corrected~ed., 1978.

\bibitem{RN189}
A.~Alexandrescu, {\em Modern C++ Design: Generic Programming and Design
  Patterns Applied}.
\newblock Upper Saddle River, NJ: Addison-Wesley, 2001.

\bibitem{RN221}
B.~Khoussainov and N.~Khoussainova, {\em Lectures on Discrete Mathematics for
  Computer Science}.
\newblock Singapore: World Scientific Publishing, 2012.

\bibitem{DelFabbro.2022}
O.~Del~Fabbro and P.~Christen, ``{Philosophy-Guided Modelling and
  Implementation of Adaptation and Control in Complex Systems},'' in {\em IEEE
  World Congress On Computational Intelligence (IEEE WCCI)}, 2022.
\newblock arXiv:2009.00110 [cs.NE].

\bibitem{RN204}
J.~von Neumann, ``The general and logical theory of automata,'' in {\em
  Cerebral Mechanisms in Behavior: The Hixon Symposium} (L.~A. Jeffress, ed.),
  New York, NY: John Wiley \& Sons, 1951.

\bibitem{RN205}
J.~Von~Neumann, {\em {Theory of Self-Reproducing Automata (A. W. Burks, ed.)}}.
\newblock Urbana and London: University of Illinois Press, 1966.

\bibitem{Burks.1970}
A.~W. Burks, {\em {Essays on Cellular Automata}}.
\newblock Urbana, Chicago, and London: University of Illinois Press, 1970.

\bibitem{RN206}
M.~Gardner, ``Mathematical games - the fantastic combinations of {John
  Conway's} new solitaire game ``life'','' {\em Scientific American}, vol.~223,
  no.~4, pp.~120--123, 1970.

\bibitem{sep-cellular-automata}
F.~Berto and J.~Tagliabue, ``Cellular automata,'' in {\em The Stanford
  Encyclopedia of Philosophy} (E.~N. Zalta, ed.), Metaphysics Research Lab,
  Stanford University, fall 2017~ed., 2017.

\bibitem{RN96}
M.~Cook, ``Universality in elementary cellular automata,'' {\em Complex
  Systems}, vol.~15, no.~1, pp.~1--40, 2004.

\bibitem{RN16}
H.~Gutowitz, {\em Cellular Automata: Theory and Experiment}.
\newblock Cambridge, MA: MIT Press, 1991.

\bibitem{RN35}
A.~Ilachinski, {\em Cellular Automata: A Discrete Universe}.
\newblock Singapore: World Scientific Publishing, 2001.

\bibitem{ann_zoo}
F.~Van~Veen and S.~Leijnen, ``{The Neural Network Zoo}.''
  https://www.asimovinstitute.org/neural-network-zoo/, 2019.
\newblock [Accessed 10 February 2020].

\bibitem{RN344}
F.~C. Richards, T.~P. Meyer, and N.~H. Packard, ``Extracting cellular automaton
  rules directly from experimental-data,'' {\em Physica D}, vol.~45, no.~1-3,
  pp.~189--202, 1990.

\bibitem{Whitehead.1967}
A.~N. Whitehead, {\em {Science and the Modern World}}.
\newblock New York, NY: Free Press, 1967.

\end{thebibliography}


\end{document}